\documentclass{article}

    \usepackage[preprint]{neurips_2024}
\makeatletter
\def\@noticestring{}
\makeatother

\usepackage[utf8]{inputenc} 
\usepackage{CJKutf8}        
\usepackage[T1]{fontenc}    
\usepackage{hyperref}       
\usepackage{url}            
\usepackage{booktabs}       
\usepackage{multirow}       
\usepackage{graphicx}       
\usepackage{amsmath}        
\usepackage{amsfonts}       
\usepackage{nicefrac}       
\usepackage{microtype}      
\usepackage{xcolor}         
\usepackage{xspace}         
\usepackage{svg}
\usepackage{booktabs}
\usepackage{multirow}
\usepackage{makecell}
\usepackage{rotating}   
\usepackage{adjustbox}
\usepackage{float}
\usepackage{threeparttable}

    \usepackage{wrapfig}
    \usepackage{booktabs}
    \usepackage{makecell}
    \usepackage{caption}

\newcommand{\github}{\raisebox{-1.5pt}{\includegraphics[height=1.05em]{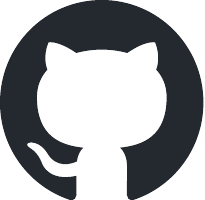}}\xspace}
\newcommand{\huggingface}{\raisebox{-1.5pt}{\includegraphics[height=1.05em]{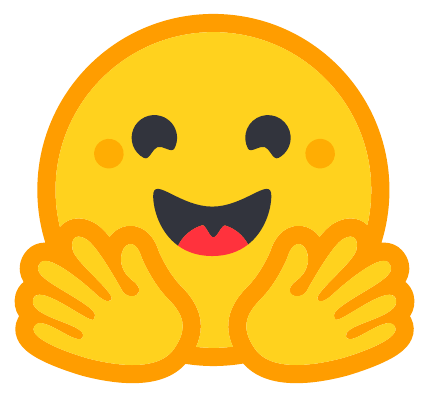}}\xspace}
\newcommand{\demo}{\raisebox{-1.5pt}{\includegraphics[height=1.05em]{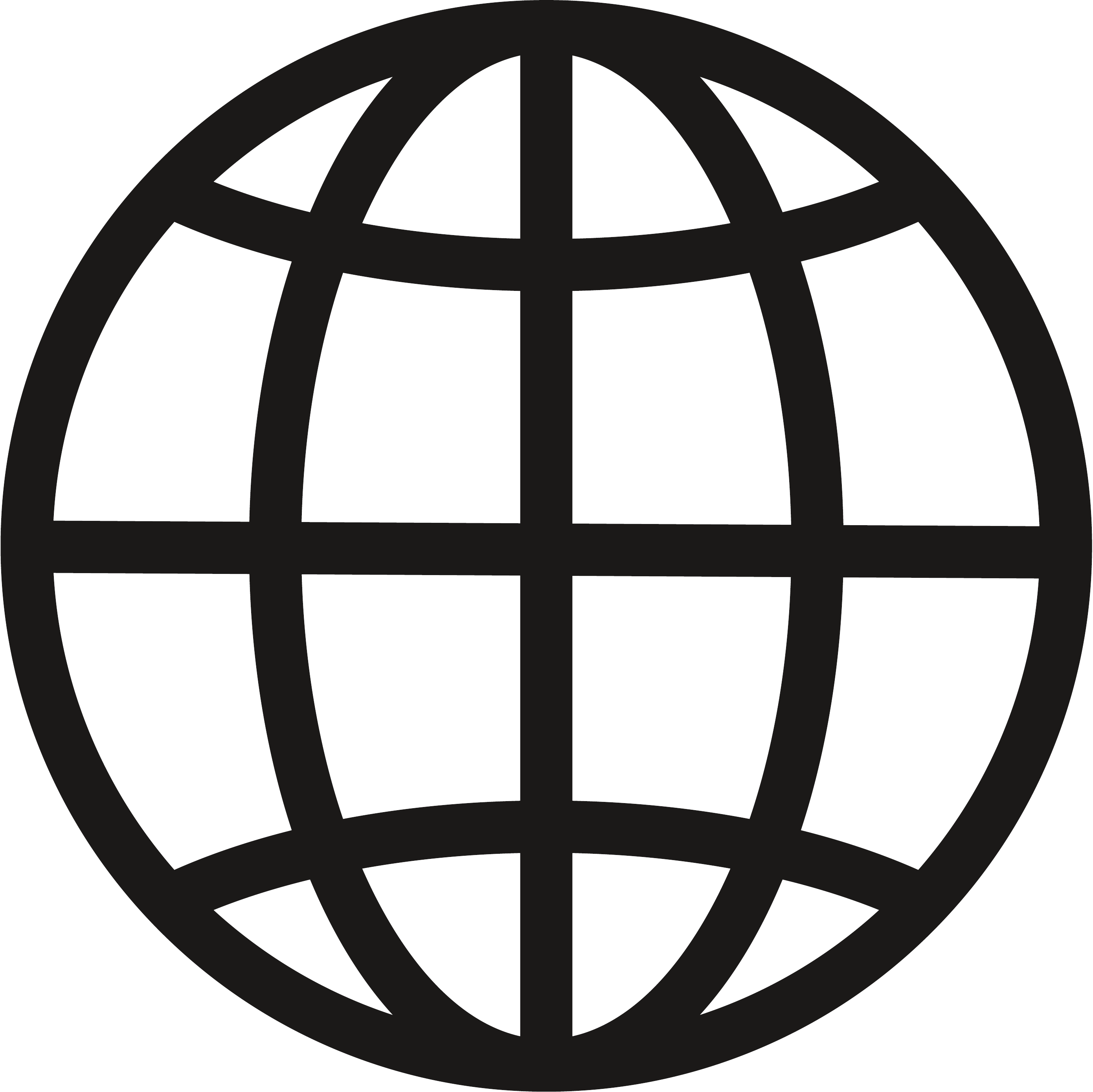}}\xspace}

\title{MiniCPM-o 4.5: Towards Real-Time Full-Duplex Omni-Modal Interaction}

\author{%
\begingroup
\newcommand{\sep}{\hspace{0.66em}}%
\centering
\hspace*{-0.7em}\begin{minipage}{\textwidth}
\centering
\footnotesize
\makebox[\textwidth][c]{Junbo Cui\sep Bokai Xu\sep Chongyi Wang\sep Tianyu Yu\sep Weiyue Sun\sep Yingjing Xu\sep Tianran Wang}\\[0.2em]
\makebox[\textwidth][c]{Zhihui He\sep Wenshuo Ma\sep Tianchi Cai\sep Jiancheng Gui\sep Luoyuan Zhang\sep Xian Sun\sep Fuwei Huang}\\[0.2em]
\makebox[\textwidth][c]{ Moye Chen\sep Zhuo Lin\sep Hanyu Liu\sep Qingxin Gui\sep Qingzhe Han\sep Yuyang Wen\sep Huiping Liu}\\[0.2em]
\makebox[\textwidth][c]{Rongkang Wang\sep Yaqi Zhang\sep Hongliang Wei\sep Chi Chen\sep You Li\sep Kechen Fang\sep Jie Zhou}\\[0.2em]
\makebox[\textwidth][c]{Yuxuan Li\sep Guoyang Zeng\sep Chaojun Xiao\sep Yankai Lin\sep Xu Han}\\[0.2em]
\makebox[\textwidth][c]{Maosong Sun\thanks{Corresponding authors.}\sep Zhiyuan Liu\footnotemark[1]\sep Yuan Yao\footnotemark[1]}\\[2mm]
\small MiniCPM-o Team, OpenBMB \\[1.5mm]
\normalsize
\demo~\href{https://minicpmo45.modelbest.cn/}{MiniCPM-o 4.5 Demo}
\quad
\huggingface~\href{https://huggingface.co/openbmb/MiniCPM-o-4_5}{MiniCPM-o 4.5 Model}
\quad
\github~\href{https://github.com/OpenBMB/MiniCPM-o}{MiniCPM-o 4.5 Code}
\end{minipage}%
\vspace{-6mm}%
\endgroup
}

\begin{document}

\maketitle
\pagestyle{plain}

\begin{figure}[h]
    \centering
    \includegraphics[width=0.85\linewidth]{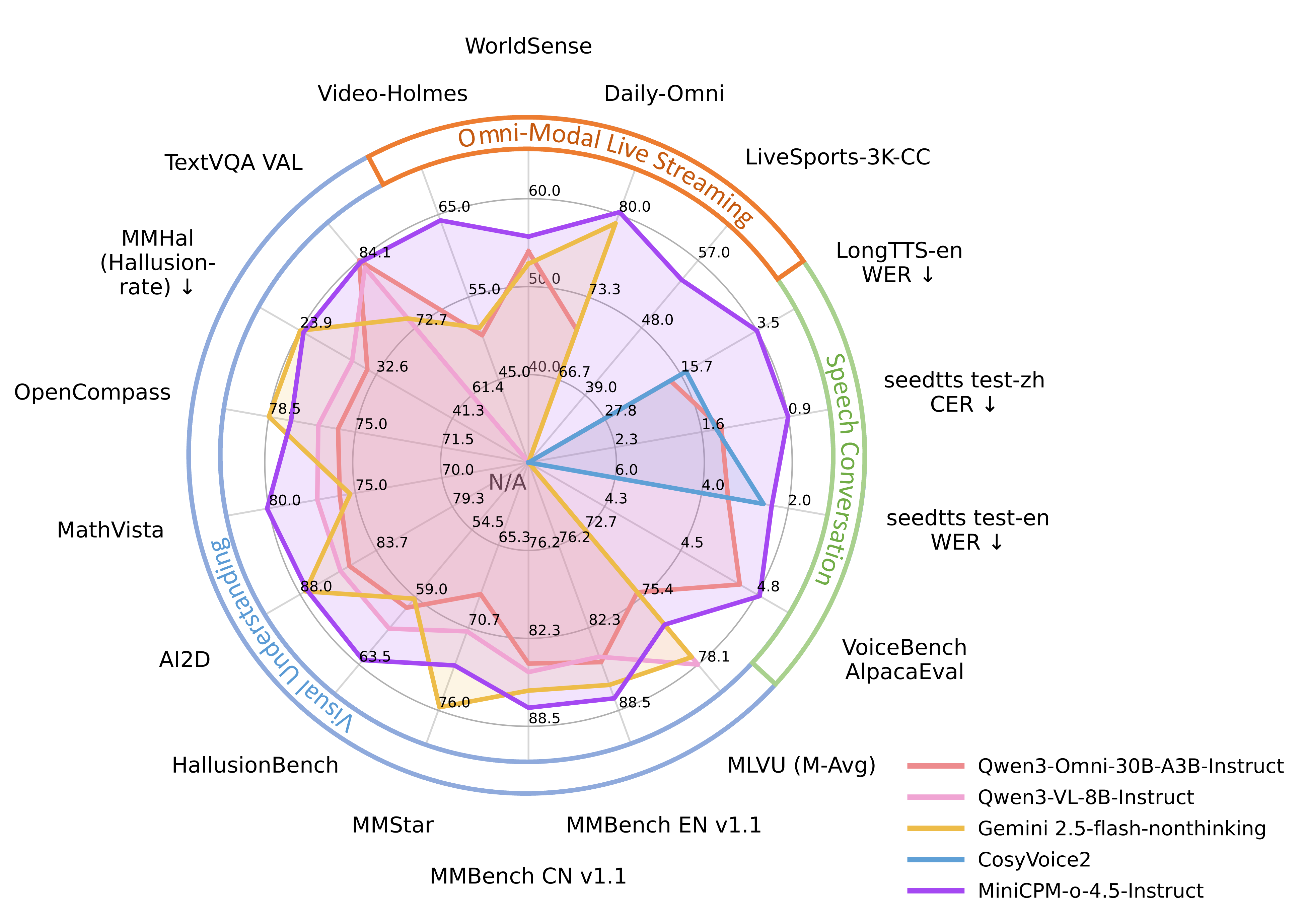}
    \caption{\textbf{Evaluation results on diverse capabilities.} MiniCPM-o 4.5 achieves state-of-the-art open-source vision-language performance at its scale, approaching Gemini 2.5 Flash. It also surpasses Qwen3-Omni-30B-A3B in omni-modal capabilities and speech generation quality.}
    \label{fig:placeholder}
\end{figure}

\begin{abstract}

Recent progress in multimodal large language models (MLLMs) has brought
AI capabilities from static offline data processing to real-time streaming interaction, yet they still remain far from human-level multimodal interaction. The key bottlenecks are no longer modality coverage or latency alone, but the interaction paradigm itself.
First, perception and response are still separated into alternating phases, preventing models from incorporating new inputs for timely adjustment during generation. Second, most current models remain reactive, responding only to explicit user requests instead of acting proactively in the evolving multimodal environment.
We present \textbf{MiniCPM-o 4.5}, our latest effort towards human-like multimodal interaction, which mitigates these gaps by real-time full-duplex omni-modal interaction. 
It can see, listen, and speak simultaneously in real-time, while also exhibiting proactive behaviors such as issuing reminders or comments based on its continuous understanding of the live scene. 
The key technique behind MiniCPM-o 4.5 is \textbf{Omni-Flow}, a unified streaming framework that aligns omni-modal inputs and outputs along a shared temporal axis. This formulation converts conventional turn-based interaction into a full-duplex, time-aligned process, enabling simultaneous perception and response and allowing proactive behavior to arise within the same framework. 
With a total of 9B parameters, MiniCPM-o 4.5 approaches Gemini 2.5 Flash in vision-language capabilities, delivering state-of-the-art open-source performance at its scale.
It also surpasses Qwen3-Omni-30B-A3B in omni-modal understanding and delivers better speech generation, with significantly higher computation efficiency.
Driven by its efficient architecture design and inference optimization, the model can perform real-time full-duplex omni-modal interaction on edge devices with less than 12GB RAM cost.
More importantly, MiniCPM-o 4.5 can be viewed as a representative example of a promising trend (Figure~\ref{fig:model_trend}): Multimodal foundation models are shipping towards human-like interactive paradigms, poised to engage with the dynamic omni-modal world in the near future.
\end{abstract}

\begin{figure}[!hb]
    \centering
    \includegraphics[width=0.94\linewidth]{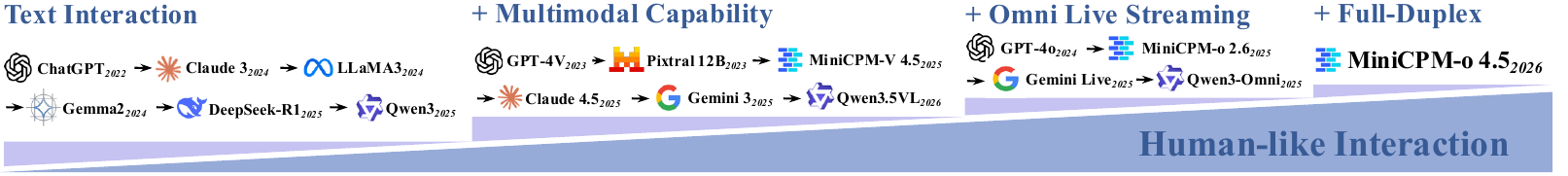}
    \caption{\textbf{Evolution of AI interaction paradigms}. AI interaction have progressed from text-only to multimodal understanding and omni live streaming. MiniCPM-o 4.5 advances this trajectory toward more human-like full-duplex interaction by enabling simultaneous perception and response.}
    \label{fig:model_trend}
\end{figure}

\section{Introduction}
\label{sec:intro}

\begin{figure}[htbp]
    \centering
    \includegraphics[width=0.99\linewidth]{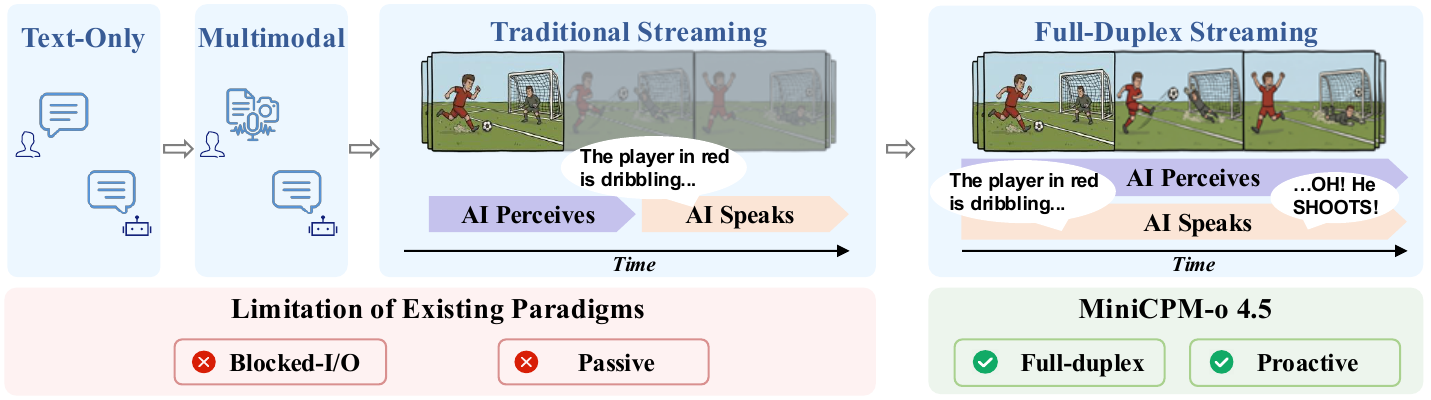}
    \caption{\textbf{From turn-based interaction to full-duplex streaming}. Existing interaction paradigms separate perception and response as alternating phases, leading to blocked information flow and passive behavior. In contrast, MiniCPM-o 4.5 continuously perceives incoming multimodal streams while speaking, allowing the model to update its response in real time and act proactively.}
    \label{fig:teaser}
\end{figure}

Progress in multimodal large language models (MLLMs) has enabled increasingly rich interaction over images, speech, video, and text, bringing AI systems closer to more natural forms of communication~\cite{yao2024minicpm,yu2025minicpmv45cookingefficient,bai2025qwen25vltechnicalreport,bai2025qwen3vltechnicalreport} (Figure~\ref{fig:model_trend}). 
The main challenge towards human-like interaction now is no longer modality coverage or response latency alone, but the underlying interaction paradigm. In current models, perception and response are still confined to alternating phases, making it difficult to continuously incorporate newly arriving information for timely adjustment during generation, as shown in Figure~\ref{fig:teaser}. Moreover, model behaviors remain strictly request-driven, rather than being proactively initiated from the evolving multimodal environment.

Tackling this challenge requires moving beyond turn-based passive response generation to continuous and proactive interaction. First, perception and response should remain continuously coupled in token-level over time, so that listening, watching, speaking, and writing can proceed in parallel instead of being forced into a serialized pipeline. 
Second, interaction should be more context-driven rather than purely reactive. Instead of waiting for explicit user triggers, a more human-like model should be able to initiate appropriate behaviors from ongoing context, such as delivering real-time scene description or offering reminders. This is particularly important in long-horizon assistance and ambient interaction.

We present MiniCPM-o 4.5, our latest effort towards human-like multimodal interaction. It can see, listen, and speak simultaneously in real-time, while also exhibiting proactive behaviors such as issuing reminders or comments based on its continuous understanding of the live scene. The key technique behind this model is 
Omni-Flow, a unified streaming framework that aligns multimodal inputs and outputs along a shared temporal axis. Rather than treating interaction as a sequence of distinct turns, Omni-Flow formulates interaction as a continuous full-duplex process, in which perception and response unfold in parallel and proactive behaviors can emerge from ongoing context within the same interaction loop. To fully exploit the rich omni-modal knowledge during training, MiniCPM-o 4.5 is built on an end-to-end multimodal architecture featuring token-level continuous connections. We also devise a time-aligned interleaving speech generation strategy, ensuring output speech is tightly aligned with the concurrent environment context.

For better compatibility with existing infrastructure and applications, MiniCPM-o 4.5 also supports traditional turn-based interaction and can be flexibly switched between the full-duplex omni-modal streaming mode and the traditional usage mode (like MiniCPM-o 2.6 and MiniCPM-V 4.5, with upgraded performance).
Extensive evaluation shows that the model achieves leading vision-language and omni-modal capabilities.  With a total of 9B parameters, it approaches Gemini 2.5 Flash in vision-language capabilities, delivering state-of-the-art open-source performance at its scale. It surpasses Qwen3-Omni-30B-A3B in omni-modal understanding and also delivers higher quality speech generation.
Taking advantage of its end-to-end continuous connections, MiniCPM-o 4.5 can accept multimodal system prompts that contain both text and reference audio, thus supporting advanced speech generation capabilities such as voice cloning.
Moreover, MiniCPM-o 4.5 retains the strong visual strengths of the MiniCPM family, including robust OCR, low hallucination, and multilingual support.

Our contributions are three-fold:(1) We present \textbf{MiniCPM-o 4.5} 9B, the first full-duplex omni-modal LLM. It can run efficiently on edge devices with less than 12GB RAM. 
(2) Extensive evaluations show that MiniCPM-o 4.5 approaches Gemini 2.5 Flash in vision-language capabilities and achieves state-of-the-art open-source performance at its scale. It also surpasses Qwen3-Omni-30B-A3B in omni-modal understanding and speech generation quality, with significantly higher computational efficiency.
(3) We identify continuous full-duplex and proactive multimodal interaction as a key step toward more human-like interactive intelligence, and propose the Omni-Flow framework, which aligns multimodal inputs and outputs along a shared temporal axis for full-duplex interaction modeling.

\section{End-to-End Omni-Modal Architecture}
\label{sec:architecture}

\begin{figure}
    \centering
    \includegraphics[width=\textwidth]{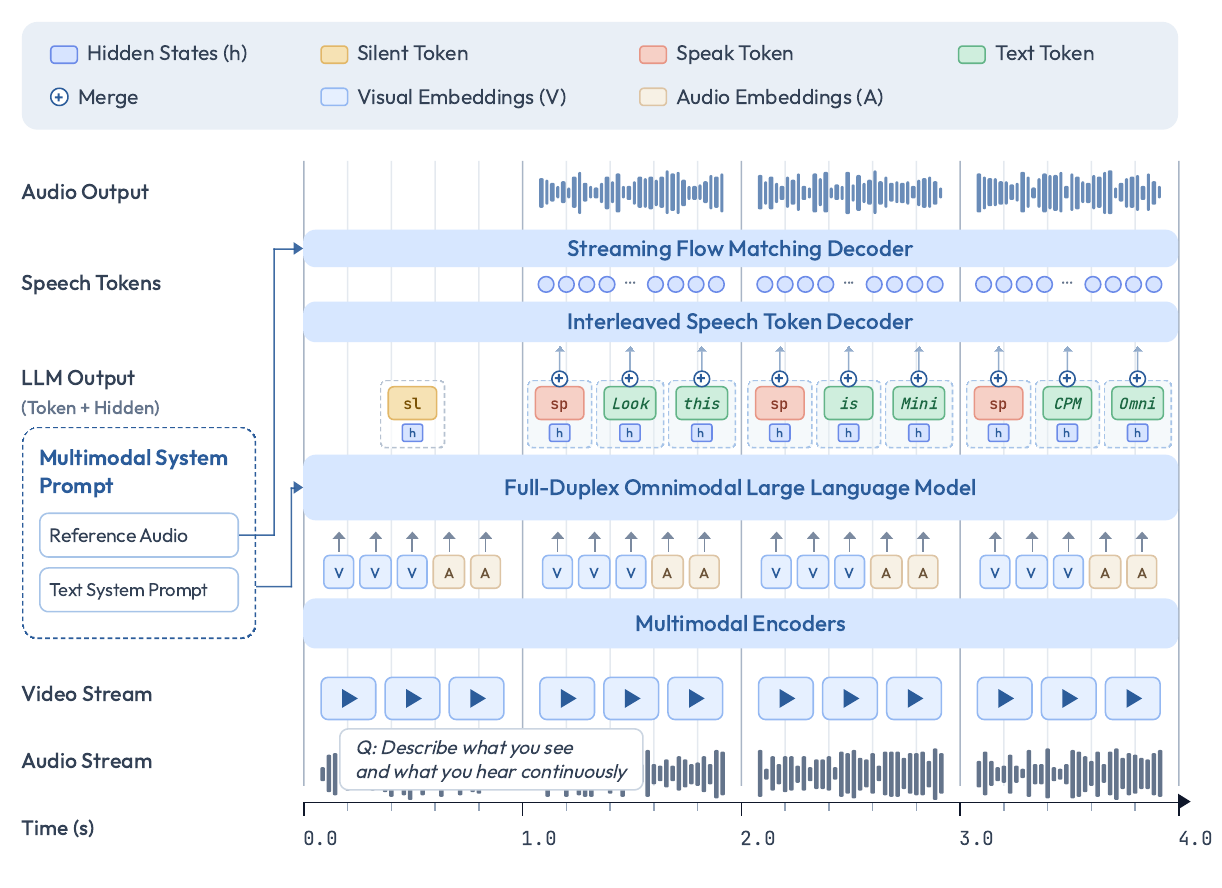}
    \caption{\textbf{End-to-end omni-modal architecture of MiniCPM-o 4.5.}
Modality encoders, the LLM backbone, and speech decoders are connected through token-level hidden states in an end-to-end trainable architecture, with multimodal input and output streams aligned on a shared millisecond-level timeline for full-duplex streaming interaction.}
    \label{fig:overview}
\end{figure}

MiniCPM-o 4.5 is built on an end-to-end omni-modal architecture that supports both full-duplex interaction under Omni-Flow and conventional turn-based inference. As illustrated in Figure~\ref{fig:overview}, it comprises three main components: (1) \textbf{multimodal encoders} that process visual and audio inputs in an streaming manner; (2) an \textbf{LLM backbone} that performs omni-modal understanding and text generation; and (3) \textbf{speech decoders}, including an interleaved speech token decoder that autoregressively generates discrete speech tokens and a streaming flow-matching decoder that converts speech tokens into audio waveforms.
All learnable components—from multimodal encoders through the LLM backbone to the speech token decoder, totaling approximately 9B parameters—are differentiably connected in token-level, enabling end-to-end gradient propagation and joint optimization across modalities during training.
Detailed architectural configurations are provided in Appendix~\ref{app:config}.

\textbf{Visual Encoding}. MiniCPM-o 4.5 adopts the LLaVA-UHD~\cite{guo2024llava} image partitioning strategy to encode any aspect high-resolution images and improve compression rate with a resampler module~\cite{yao2024minicpm}. We adopt a max resolution of 448$\times$448 for the full-duplex streaming mode and otherwise 2240$\times$2240. Specifically, each image is first divided into slices, and each slice is then encoded into 1024 tokens by a SigLIP ViT~\cite{zhai2023sigmoid} (0.4B) and compressed into 64 tokens by the resampler module. This yields a 16$\times$ token compression ratio, which is higher than the common 4$\times$ compression~\cite{xu2025qwen3omnitechnicalreport, bai2025qwen25vltechnicalreport, bai2025qwen3vltechnicalreport}, enabling substantially more efficient visual processing.

\textbf{Audio Encoding}. A Whisper Medium~\cite{radford23a} encoder (0.3B) encodes input audio in a chunk-based streaming fashion~\cite{yao2021wenet}, producing 50 feature tokens per second. We then use a two-layer MLP projector to conduct a 5$\times$ temporal compression, resulting in 10 audio tokens per second for the LLM backbone, reducing the token budget. 

\textbf{Text Decoding}. The LLM backbone (Qwen3-8B~\cite{qwen3technicalreport}) generates text outputs and hidden states for speech generation. Since the LLM backbone only generate tokens in text domain, it requires just 3-4 decoding steps per second (i.e., human speech speed) during real-time full-duplex interaction. When backbones are instead required to directly generate speech tokens (typically about 25 tokens per second), as in recent works~\cite{xie2024miniomnilanguagemodelshear, stepfun2025stepaudio2}, the efficiency can be significantly impeded, and the core language capabilities also tend to degrade~\cite{hsiao2025analyzing,xu2025qwen25omnitechnicalreport}. Our design avoids this by delegating speech token production to lightweight speech decoders described below.

\textbf{Speech Token Generation}. Speech generation demands not only correct pronunciation but also prosody and style shaped by context and instructions. We address this by leveraging the contextual understanding capability of the LLM backbone.
For each text token passed to the lightweight Llama speech token decoder (${\sim}$0.3B), we sum its LLM backbone hidden states (reshaped by an MLP layer) and its speech decoder for further S3~\cite{du2024cosyvoice} token generation. 
With prosodic decisions pre-encoded by the LLM backbone, the small speech decoder can devote its capacity to speech modeling. Moreover, input text tokens and output speech tokens are interleaved in a time-aligned manner to ensure output speech tightly couples with the concurrent environment context as detailed in Section~\ref{TAIL}.

\textbf{Waveform Synthesis}. A streaming flow-matching decoder~\cite{du2024cosyvoice2,stepfun2025stepaudio2} converts generated S3 speech tokens into audio waveforms, based on the reference audio in the multimodal system prompt.

\section{Omni-Flow}

In existing interaction paradigms, perception and response are confined to alternating phases, resulting in the blocked I/O and passive responding problem as illustrated in Figure~\ref{fig:teaser}. To enable models to perceive and speak simultaneously, we propose the Omni-Flow framework that coordinates omni-modal input and output streams with a shared temporal axis.
Inspired by the time-division multiplexing technique, Omni-Flow partitions the continuous interaction into fine-grained time windows of duration~$t$. Within each window, the model incorporates newly arrived signals while producing the next output, converting conventional turn-taking into a stream of time-local updates as shown in Figure~\ref{fig:overview}. As $t$ becomes sufficiently small, perception and response become tightly coupled in time, naturally approximating full-duplex behavior.

\subsection{Time-Aligned Streams}

We identify three time-aligned streams in the interaction: \textbf{env-visual}, which carries live visual observations of the environment; \textbf{env-audio}, which carries the acoustic scene, including user speech when present; \textbf{out-stream}, which represents the assistant's text and speech outputs.
Under this view, user requests are no longer treated as a privileged conversational role, but instead become part of the continuously observed world state, entering primarily through env-audio. Likewise, the model does not rely on explicit requests as the trigger before responding. Instead, the out-stream evolves coupled to ongoing perception. The model is therefore situated in an always-on multimodal environment, where it must determine not only \emph{what} to output, but also \emph{whether} and \emph{when} to output on its own.

\subsection{Unified Serialization}

Given these streams, we organize them into a unified sequence that can be passed to a standard causal language model.
For the $k_\text{th}$ time chunk, inputs from env-visual and env-audio are encoded into visual token sequence $\mathbf{v}^k$ and audio token sequence $\mathbf{a}^k$, while updates in out-stream are represented as an output token sequence $\mathbf{o}^k$. When no output should be produced, $\mathbf{o}^k$ contains only a special \texttt{[listen]} token.
We group these time-aligned tokens into
$
\mathbf{g}_k = [\mathbf{v}^k;\mathbf{a}^k;\mathbf{o}^k],
$
and serialize the interaction by concatenating consecutive groups into a single sequence. 
Within each chunk, the model first processes newly arrived perceptual tokens and then generates output tokens, so that every output is conditioned on the most recent observation. Reducing the chunk size $t$ increases the rate at which the model refreshes its perception, keeping it more closely aligned with the evolving environment. Since the model determines whether to output in each time window, it naturally supports proactive behavior and reduces the reliance on external VAD~\cite{vad_sohn1999statistical} modules.

\subsection{Design Tradeoffs}

Omni-Flow introduces several design choices that directly affect the stability and responsiveness of the model. We therefore conduct ablations along three dimensions: temporal granularity, boundary explicitness, and control formulation. Temporal granularity specifies the duration of each time chunk ($1.0$\,s, $0.2$\,s, or $0.1$\,s). Boundary explicitness specifies whether consecutive groups are separated by explicit special tokens or not. Control formulation specifies how the model decides whether to speak: in the Listen-Speak (LS) formulation, the model first predicts a binary \texttt{listen}/\texttt{speak} control token before content generation; in the Listen-Text (LT) formulation, the model directly predicts either \texttt{[listen]} or normal text tokens in a shared output space. Results are shown in Table~\ref{tab:duplex_ablation}.

\begin{table}[h]
\centering
\caption{Ablation of full-duplex design choices.}
\label{tab:duplex_ablation}
\resizebox{\linewidth}{!}{
\begin{tabular}{ccc|ccccc}
\toprule
\textbf{Chunk Size} & \textbf{Boundary} & \textbf{Control} & \textbf{AdvBench} & \textbf{AlpacaEval} & \textbf{IFEval} & \textbf{SDQA} & \textbf{MMLU} \\
\midrule
\textbf{1.0\,s} & \textbf{Explicit} & \textbf{LS} & \textbf{0.98} & 3.56 & \textbf{0.29} & \textbf{0.36} & \textbf{0.65} \\
1.0\,s & Explicit & LT & 0.92 & \textbf{3.60} & 0.24 & 0.35 & 0.56 \\
1.0\,s & Implicit & LT & 0.96 & 3.31 & 0.22 & 0.28 & 0.45 \\
0.2\,s & Explicit & LS & 0.81 & 1.22 & 0.10 & 0.09 & 0.45 \\
0.1\,s & Explicit & LS & 0.67 & 2.40 & 0.10 & 0.13 & 0.32 \\
\bottomrule
\end{tabular}
}
\end{table}

\textbf{Temporal granularity governs the central latency-capacity tradeoff}. Reducing the chunk size improves temporal responsiveness, but also leaves less modeling budget within each chunk for control and generation. When chunks become too short, the model no longer has sufficient information for each time window to make stable decisions and produce coherent outputs, leading to substantial degradation. In our setting, a chunk size of $1.0$\,s provides the best balance.

\textbf{Boundary explicitness is consistently beneficial}. Explicitly marking the boundary between groups performs better. This suggests that distinguishing newly observed inputs from newly generated outputs is a nontrivial problem, and making this structure explicit can reduce the burden on the model.

\textbf{Separating interaction control from content generation leads to more stable modeling}. 
LS outperforms LT, indicating that deciding \emph{whether} to speak should be decoupled from deciding \emph{what} to say, and entangling both in a single prediction step makes full-duplex interaction harder to learn.

\subsection{Time-Aligned Interleaving for Timely Speech Generation}
\label{TAIL}

Omni-Flow represents model outputs as a stream that evolves together with incoming inputs. However, maintaining temporal alignment between the spoken output and the latest observed context remains nontrivial. The difficulty comes from the mismatch between text generation time and speech playback time: if the text generated within an $m$-second interval takes much longer than $m$ seconds to vocalize, the speech stream will progressively lag behind the model's evolving state. As a result, the audio heard at a given moment may correspond to text generated much earlier, making the response temporally stale with respect to the ongoing interaction. This issue is further complicated by the fact that the vocalization duration of each text token is variable and context-dependent.

\begin{figure}[htbp]

    \centering

    \includegraphics[width=\textwidth]{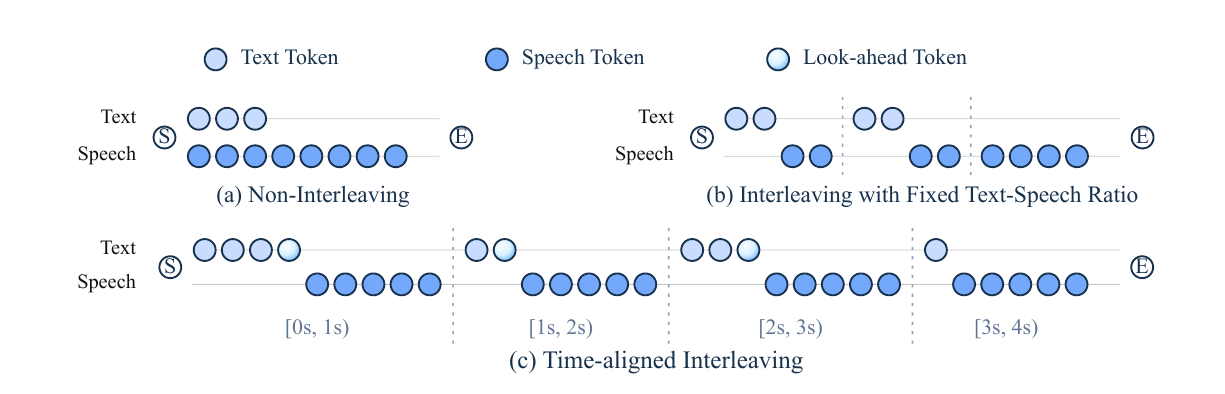}

    \caption{\textbf{Comparison of streaming speech generation strategies}. Existing methods either (a) maintain a large text lead or (b) rely on a fixed text-speech ratio, making the spoken content lag behind the evolving environment. We propose Time-Aligned Interleaving (TAIL), which adaptively interleaves text and speech so that the text generated in each time chunk corresponds to approximately the same duration of speech playback.}

    \label{fig:TAIL}

\end{figure}

Existing streaming speech generation methods~\cite{xie2024miniomnilanguagemodelshear,xu2025qwen3omnitechnicalreport,xu2025qwen3,du2024cosyvoice2} typically adopt one of two strategies shown in Figure~\ref{fig:TAIL} (a) and (b). Some methods first generate a relatively long span of text and then synthesize speech from it. Others interleave text and speech using a fixed text-to-speech token ratio. While both strategies can produce high-quality speech, they do not explicitly align the generated speech with the interaction timeline. The former allows text to run far ahead of playback, while the latter assumes a nearly fixed correspondence between text tokens and speech duration. In full-duplex interaction, both designs can cause the model to keep speaking content that is stale and not aligned with the concurrent environment. 

To address this, we propose \textbf{Time-Aligned Interleaving (TAIL)}, a chunk-wise speech generation strategy that adaptively controls how much text to generate at each step. Rather than matching each chunk independently to a fixed speech duration, TAIL considers the accumulated playback progress over the entire interaction. At the $k_\text{th}$ chunk, the model adjusts the amount of text to generate so that, after vocalizing the newly generated content, the speech stream approaches the current time boundary $kt$. If previous chunks have already introduced a slight playback delay, the model can adaptively generate fewer text tokens in the current chunk to let speech catch up. In this way, TAIL keeps the spoken response close to the model's latest state instead of allowing text to run far ahead of audio.

We construct TAIL supervision from full-duplex streaming training data by collecting the start and end times of each text token. Tokens whose start times fall into $[(k-1)t, kt)$, together with their corresponding speech tokens, are assigned to the $k_\text{th}$ Omni-Flow chunk. This format teaches the model to learn a history-dependent interleaving pattern, where the number of text tokens in each chunk can vary according to the accumulated playback alignment.

\textbf{Look Ahead Speech Generation.} Speech generation may still require a limited future text context. For example, the pronunciation of ``the'' depends on the following word, as in ``the apple'' versus ``the car''. TAIL therefore uses a bounded look-ahead mechanism: the speech tokens of the last few text tokens in chunk $k$ are deferred to chunk $k+1$, while the remaining tokens are spoken in chunk $k$. This provides local context for pronunciation and prosody without letting the text stream run substantially ahead of playback. As a result, TAIL preserves the time-aligned structure of Omni-Flow while enabling continuous and timely speech generation.
\section{Data}
\label{sec:data}

\subsection{Speech Data}

We collect large-scale natural speech data for broad capability coverage and high-quality dialog data for controllable natural speech generation.

\textbf{Large-scale Natural Speech Data}. We process millions of hours of unlabeled speech data collected from diverse sources through a pipeline integrating multiple open-source components~\cite{silero_vad_2024,radford2022robustspeechrecognitionlargescale,gao2023paraformerfastaccurateparallel,han2024leveragingselfsupervisedlearningspeaker,défossez2021musicsourceseparationwaveform}, yielding training sets for zero-shot TTS, ASR, and multi-turn multi-speaker dialogue. This diverse corpus encompasses a broad range of different speakers, accents, and conversational patterns. 

\textbf{Spoken Dialog Data}. We first use a text-based LLM to generate colloquial, instruction-following dialogue from diverse seed queries. A subset of these dialogues is then re-recorded by professional voice actors under studio conditions. 
In the recording sessions, voice actors deliver in a conversational style rather than reading scripts verbatim, balancing structured content with improvised expression while varying emotion, speaking rate, and emphasis under a consistent vocal identity. The resulting corpus covers instruction-following TTS, question answering, and multi-turn natural dialogue.

\subsection{Vision-Language Data}

We introduce the vision-language data of MiniCPM-o 4.5 in this section. Building upon the data system of MiniCPM-V 4.5, we further expand the scale and improve the quality to cover broader task types and real-world scenarios.

\textbf{High-Quality Knowledge and Alignment Data.} We update the generator model used in the CapsFusion \cite{yu2024capsfusion} pipeline to synthesize more informative image captions, and further refine our filtering process by improving image-text relevance estimation.

\textbf{Complex Document and OCR Data}. To better utilize document knowledge, we extend the unified document knowledge and OCR learning approach of MiniCPM-V 4.5 with a relevance-aware masking strategy. Specifically, instead of randomly masking text regions, we prioritize regions that are more relevant to figures and charts in document images. This encourages the model to focus more on visually grounded content, while reducing the proportion of training cases that can be solved primarily from textual context alone.

\textbf{Real-World Scenarios Data.} Capturing the nuances of practical user interactions is a core focus of our data curation. We introduce more natural and diverse query patterns. We significantly improve the depth and readability of model responses by rewriting short, direct-answer samples into detailed, chain-of-thought-style rationales. In addition, a reward-model-based filtering pipeline is applied to ensure overall data quality and alignment with human preferences.

\textbf{Dense Video Perception Data.} To strengthen the model's video perception and cross-frame reasoning abilities, we construct a dense video captioning dataset which provides continuous, fine-grained descriptions of temporal events, human actions, and complex scene transitions.

\textbf{Text-only Data.} We also incorporate high-quality text-only instruction data from the MiniCPM 4.1 \cite{minicpm4} post-training data set to maintain robust linguistic capabilities.

\subsection{Omni-Modal Full-Duplex Data}

Our omni-modal full-duplex data includes both large-scale web data and a smaller set of high-quality instruction samples. Each training sample contains the full visual input, audio input, output text and output speech, where each piece of information is tagged with a time index.

\textbf{Large-scale Web Audio-video Data}. We collect a large scale of web audio-video data to provide broad coverage of real-world full-duplex scenarios. Segments dominated by single-speaker speech or that have weak audio-visual relevance are filtered out. To further improve quality, we apply OCR-based subtitle removal~\cite{cui2025paddleocr30technicalreport}, talking-head detection~\cite{chen2025livecclearningvideollm}, and filtering over ASR-derived transcripts, reducing misleading shortcuts and low-information or noisy segments.

\textbf{Full-Duplex Task Data}. To support target full-duplex capabilities that require more precise interaction, we manually construct multiple scenarios and annotate corresponding instruction-following data. Based on these high-quality task samples, MiniCPM-o 4.5 supports advanced capabilities like continuous scene description and proactive reminding.

\section{Training}
\label{sec:training}

In this section, we present the overall training pipeline for MiniCPM-o 4.5. One of the key challenges in advancing omni-modal capabilities is to retain the fundamental advantages of individual modalities while supporting efficient and seamless generalization across modalities. To this end, we design a carefully staged pipeline to progressively integrate speech into the multimodal system in a smooth and stable manner.
Based on a pretraining checkpoint of MiniCPM-V 4.5.
The pipeline first conducts speech pretraining to establish foundational audio understanding and speech generation capabilities. We then perform joint pretraining to construct unified cross-modal representations. Supervised fine-tuning is further employed to enable natural instruction following and high-quality interactions across text, speech, image, and video. Finally, we apply reinforcement learning to further improve reasoning abilities and mitigate hallucinations.

\subsection{Speech Pretraining}

MiniCPM-o 4.5 is initialized with a pretrained Whisper encoder and the pretraining checkpoint of MiniCPM-V 4.5, together with randomly initialized speech-related modules, including an audio projector, an LLM-to-speech projector, and a speech decoder. To preserve the backbone's visual and linguistic capabilities, we freeze the pretrained components and update only newly added modules. This stage aligns Whisper features with the LLM hidden space and trains the speech decoder to transform LLM backbone hidden states into semantically and prosodically grounded speech tokens.

\subsection{Joint Pretraining}

In the second stage, we unfreeze all parameters and conduct joint pretraining on a balanced mixture of vision-language, speech, and omni-modal data. To stabilize optimization, we assign different modality combinations to different data-parallel ranks, ensuring a fixed data ratio at every training step. Besides conventional turn-based samples, the mixture includes proactive and full-duplex interaction data, where text tokens are aligned with speech and visual signals on a shared timeline. Trained with a unified next-token prediction objective, the model acquires real-time omni-modal interaction capabilities while maintaining its foundational visual understanding.

\subsection{Joint Supervised Fine-Tuning}

The joint supervised fine-tuning stage activates omni-modal capabilities and strengthens instruction following. It consists of two phases: large-scale instruction tuning for broad capability adaptation, followed by high-quality human-annotated tuning for fine-grained behavioral refinement. To enable flexible quality-efficiency trade-offs during inference, we augment omni-modal data with varying resolutions and frame rates, randomly setting the maximum frame resolution to $0.2$--$0.4$ megapixels and sampling the frame rate uniformly from $1$--$5$ FPS.

\subsection{Reinforcement Learning}

We further improve MiniCPM-o 4.5 with reinforcement learning. We first apply GRPO~\cite{shao2024deepseekmath} to enhance reasoning and instruction following, using answer accuracy together with auxiliary rewards such as format reward. For accuracy rewards, we combine rule-based verification with an efficient judge model~\cite{liu2025compassverifier} to improve the recall of correct responses.

To improve token efficiency, we introduce a smooth length reward adapted from Kimi-K1.5~\cite{team2025kimi}:
\begin{equation}
r_{\mathrm{len}}(i)=
\begin{cases}
s_i, & r_i=1,\\
\min(0,s_i), & r_i=0,
\end{cases}
\quad
s_i=
\left(0.5-\frac{\ell_i-\ell_{\min}}{\ell_{\max}-\ell_{\min}}\right)
\times\min\!\left(1,\frac{\ell_{\max}-\ell_{\min}}{\tau}\right).
\end{equation}
Here, \(r_i\) is the correctness indicator, and \(\ell_i,\ell_{\min},\ell_{\max}\) are computed over responses to the same prompt. The \(\min(0,s_i)\) term avoids rewarding short incorrect responses, and \(\tau\) downscales the reward when length differences are small. We also include a general reward model to improve answer quality and suppress unintended code-mixing. For convergence efficiency, we do not include the length reward for the first 480 training steps.

Finally, we apply RLAIF-V~\cite{yu2024rlaifvopensourceaifeedback} to reduce hallucinations in visual scenarios. We find that hallucination mitigation learned from image-text data transfers effectively to omni-modal full-duplex interaction, reducing hallucinations in streaming settings as well.

\section{Evaluation}
\label{sec:experiments}

In this section, we comprehensively evaluate MiniCPM-o 4.5 and other baseline models.

\subsection{Modalities and Domains}

We evaluate MiniCPM-o 4.5 across four modality capability groups: vision-language understanding, speech understanding and generation, text capability, and omni-modal streaming interaction. 
Vision-language understanding is further divided into five representative domains: STEM and general multimodal reasoning, document and OCR understanding, multi-image reasoning, hallucination, and video understanding. 
Speech evaluation covers both speech understanding and speech generation. 
Text evaluation measures whether the model preserves the language capabilities of its LLM backbone after omni-modal training. 
Omni-modal and streaming interaction evaluation covers both turn-based omni-modal understanding and full-duplex streaming interaction.

\begin{table}
  \centering
  \normalsize
  \caption{Vision-language results (instruct mode).}
  \label{tab:main-results-instruct}
  \resizebox{\textwidth}{!}{%
  \begin{tabular}{lcccc|c}
    \toprule
    \textbf{Benchmark} & \makecell{\textbf{Gemini 2.5 Flash}} & \makecell{\textbf{InternVL3.5}} & \makecell{\textbf{Qwen3-VL}} & \makecell{\textbf{Qwen3-Omni}} & \textbf{MiniCPM-o 4.5} \\
    \textbf{Size} & \textbf{-} & \textbf{8B} & \textbf{8B} & \textbf{30B-A3B} & \textbf{9B} \\
    \midrule
    \multicolumn{6}{l}{\textbf{STEM \& General}} \\
    \hspace{1.2em}OpenCompass & \textbf{78.5} & 75.8 & 76.5 & 75.7 & 77.6 \\
    \hspace{1.2em}MMBench EN v1.1 & 86.6 & 79.5 & 84.5 & 84.9 & \textbf{87.6} \\
    \hspace{1.2em}MMBench CN v1.1 & 86.0 & 80.0 & 84.7 & 84.1 & \textbf{87.2} \\
    \hspace{1.2em}MathVista & 75.3 & 78.4 & 77.2 & 75.9 & \textbf{80.1} \\
    \hspace{1.2em}MMVet & 81.4 & \textbf{83.1} & 73.7 & 74.8 & 74.4 \\
    \hspace{1.2em}MMMU & \textbf{76.3} & 73.4 & 69.6 & 69.1 & 67.6 \\
    \hspace{1.2em}MMStar & \textbf{75.8} & 69.3 & 70.9 & 68.5 & 73.1 \\
    \hspace{1.2em}AI2D & \textbf{87.7} & 84.0 & 85.7 & 85.2 & 87.6 \\
    \hspace{1.2em}MMT-Bench (val) & 70.0 & 66.7 & 60.9 & \textbf{70.4} & 69.7 \\
    \hspace{1.2em}MM-IFEval & \textbf{75.8} & 56.3 & 59.4 & 65.7 & 66.3 \\
    \midrule
    \multicolumn{6}{l}{\textbf{Document \& OCR}} \\
    \hspace{1.2em}OCRBench & 864 & 840 & \textbf{896} & 880 & 876 \\
    \hspace{1.2em}TextVQA (val) & 74.3 & 78.2 & 82.9 & \textbf{84.1} & 83.8 \\
    \hspace{1.2em}DocVQA (val) & 93.0 & 92.3 & \textbf{96.1} & 95.4 & 94.7 \\
    \hspace{1.2em}OmniDocBench (EN)$\downarrow$ & 0.214 & 0.322 & 0.255 & 0.216 & \textbf{0.109} \\
    \hspace{1.2em}OmniDocBench (CN)$\downarrow$ & 0.290 & 0.416 & 0.319 & 0.363 & \textbf{0.162} \\
    \midrule
    \multicolumn{6}{l}{\textbf{Hallucination}} \\
    \hspace{1.2em}HallusionBench & 59.1 & 54.5 & 61.1 & 59.7 & \textbf{63.2} \\
    \hspace{1.2em}MMHal-Score & \hspace{1.3mm}4.6 & \hspace{1.3mm}3.8 & \hspace{1.3mm}4.7 & \hspace{1.3mm}4.6 & \hspace{1.3mm}\textbf{4.7} \\
    \hspace{1.2em}MMHal-Hallrate$\downarrow$ & \textbf{23.9} & 34.7 & 29.9 & 31.6 & 24.3 \\
    \midrule
    \multicolumn{6}{l}{\textbf{Multi-Image}} \\
    \hspace{1.2em}Mantis-Eval & 72.8 & 70.5 & 74.2 & 78.3 & \textbf{79.7} \\
    \hspace{1.2em}MUIRBench & \textbf{74.5} & 55.8 & 64.4 & 61.9 & 72.0 \\
    \hspace{1.2em}MMSI-Bench & 12.1 & -- & 11.3 & 14.2 & \textbf{16.6} \\
    \midrule
    \multicolumn{6}{l}{\textbf{Video}} \\
    \hspace{1.2em}Video-MME (w/o subs) & \textbf{75.6} & 66.0 & 71.4 & 70.5 & 70.4 \\
    \hspace{1.2em}LVBench & \textbf{62.2} & -- & 58.0 & 50.2 & 50.9 \\
    \hspace{1.2em}MLVU (M-Avg) & 77.8 & 70.2 & \textbf{78.1} & 75.2 & 76.5 \\
    \hspace{1.2em}LongVideoBench (val) & -- & 62.1 & 66.4 & \textbf{66.9} & 66.0 \\
    \hspace{1.2em}MotionBench & -- & \textbf{62.3} & 59.5 & 61.7 & 61.4 \\
    \bottomrule
  \end{tabular}
  }
\end{table}

\textbf{Vision-Language Understanding.}
We evaluate vision-language understanding across five representative domains.
(1) \emph{STEM and general multimodal reasoning}.
For general vision-language comprehension, we include OpenCompass~\citep{2023opencompass}, MMBench V1.1~\citep{liu2024mmbench}, MMVet~\citep{yu2023mm}, and MMStar~\citep{chen2024we}, which cover diverse multimodal tasks. 
For STEM-oriented reasoning, we include MMMU~\citep{yue2024mmmu}, MathVista~\citep{lu2024mathvista}, and AI2D~\citep{kembhavi2016diagram}, covering scientific knowledge, mathematical reasoning, and diagram understanding. 
We further include MMT-Bench~\citep{ying2024mmt} and MM-IFEval~\citep{ding2025mm} to assess multitask generalization and multimodal instruction following.
(2) \emph{Document and OCR understanding}.
This domain evaluates the ability to recognize, extract, and reason over text in visually rich documents and scene images. 
We use OCRBench~\citep{liu2024ocrbench}, TextVQA~\citep{singh2019textvqa}, DocVQA~\citep{mathew2021docvqa}, and OmniDocBench~\citep{ouyang2025omnidocbenchbenchmarkingdiversepdf}, which require joint modeling of textual content, visual layout, and document structure.
(3) \emph{Multi-image understanding}.
This domain measures the ability to aggregate and compare information across multiple images. 
We adopt Mantis-Eval~\citep{jiang2024mantis}, MUIRBench~\citep{wang2024muirbench}, and MMSI-Bench~\citep{yang2025mmsi}, which evaluate cross-image reasoning, visual comparison, and multi-image information integration.
(4) \emph{Hallucination}.
This domain evaluates whether model responses remain faithful to the visual input. 
We use HallusionBench~\citep{guan2024hallusionbench} and MMHal-Bench~\citep{sun2023aligning}, which measure visual consistency and hallucination in multimodal generation.
(5) \emph{Video understanding}.
This domain evaluates spatio-temporal reasoning and motion understanding in videos. 
We use Video-MME~\citep{fu2024videomme}, LVBench~\citep{wang2024lvbench}, MLVU~\citep{zhou2025mlvu}, LongVideoBench~\citep{wu2024longvideobench}, and MotionBench~\citep{hong2024motionbench}, covering both varying video lengths.

\textbf{Speech Understanding and Generation.}
Speech evaluation covers automatic speech recognition, speech translation, audio understanding, speech question answering, and speech generation. 
For speech understanding, we evaluate on standard ASR benchmarks, including AISHELL-1~\citep{bu2017aishell}, AISHELL-2~\citep{du2018aishell2}, WenetSpeech~\citep{zhang2022wenetspeech}, LibriSpeech~\citep{panayotov2015librispeech}, GigaSpeech~\citep{chen2021gigaspeech}, and VoxPopuli~\citep{wang2021voxpopuli}; speech translation on CoVoST 2~\citep{wang2020covost2}; multi-task audio understanding on MMAU and MELD~\citep{poria2019meld}; and spoken question answering on VoiceBench~\citep{voicebench}, Speech TriviaQA~\citep{joshi2017triviaqa}, Speech Web Questions~\citep{berant2013semantic}, and Speech CMMU~\citep{li2023cmmlu}. 
For speech generation, we evaluate speech quality, intelligibility, speaker similarity, long-form generation, and emotion/style control using SeedTTS Test~\citep{anastassiou2024seedttsfamilyhighqualityversatile}, LongTTS~\citep{longtts}, Expresso~\citep{expresso}, and ESD~\citep{zhou2021esd}.

\textbf{Text Capability.}
We compare MiniCPM-o 4.5 with its language backbone, Qwen3-Instruct-8B~\citep{qwen3technicalreport}, to assess whether omni-modal training preserves core text abilities.
Our benchmark suite spans instruction following, world knowledge, multilingual understanding, reasoning, and code generation.
Specifically, we use IFEval~\citep{zhou2023instruction} for instruction following; MMLU~\citep{hendrycks2021measuring} and CMMLU~\citep{li2024cmmlu} for knowledge and multilingual understanding; BBH~\citep{suzgun2023challenging}, MATH-500~\citep{hendrycks2021math}, and GSM8K~\citep{cobbe2021training} for reasoning and mathematics; and HumanEval~\citep{chen2021evaluating} and MBPP~\citep{austin2021program} for code generation.

\textbf{Omni-modal and Streaming Interaction.}
We evaluate omni-modal understanding on benchmarks where video and audio input streams are naturally time-aligned, including Daily-Omni~\citep{zhou2025dailyomni}, WorldSense~\citep{hong2025worldsense}, Video-Holmes~\citep{cheng2025videoholmes}, JointAVBench~\citep{chao2025jointavbench}, AVUT-Human~\citep{yang2025avut}, FutureOmni~\citep{chen2026futureomni}, and Video-MME-Short with audio~\citep{fu2024videomme}. 
For full-duplex streaming, the model must continuously perceive incoming streams while producing timely responses. 
Due to the limited availability of benchmarks for real-time omni-modal full-duplex interaction, we report results on LiveSports-3K-CC~\citep{chen2025livecclearningvideollm}, an audio-free full-duplex benchmark. 
Qualitative demonstrations involving simultaneous vision, speech, and text streams are provided on our demo website.

\begin{table}
  \centering
  \normalsize
  \caption{Vision-language results (thinking mode).}
  \label{tab:main-results-thinking}
  \resizebox{\textwidth}{!}{%
  \begin{tabular}{l c >{\centering\arraybackslash}m{1.8cm} c c | c}
    \toprule
    \textbf{Benchmark} & \makecell{\textbf{Gemini 2.5 Flash}} & \textbf{GPT-5} & \makecell{\textbf{Qwen3-VL}} & \makecell{\textbf{Qwen3-Omni}} & \textbf{MiniCPM-o 4.5} \\
    \textbf{Size} & \textbf{-} & \textbf{-} & \textbf{8B} & \textbf{30B-A3B} & \textbf{9B} \\
    \midrule
    \multicolumn{6}{l}{\textbf{STEM \& General}} \\
    \hspace{1.2em}OpenCompass & \textbf{79.9} & 79.7 & 77.3 & 78.5 & 78.2 \\
    \hspace{1.2em}MMBench EN v1.1 & 87.1 & 85.5 & 85.3 & 88.2 & \textbf{89.0} \\
    \hspace{1.2em}MMBench CN v1.1 & 87.3 & 85.6 & 85.5 & \textbf{87.7} & 87.6 \\
    \hspace{1.2em}MathVista & 79.4 & \textbf{81.9} & 81.4 & 80.0 & 81.0 \\
    \hspace{1.2em}MMVet & \textbf{81.2} & 77.6 & 69.8 & 74.8 & 73.6 \\
    \hspace{1.2em}MMMU & 77.7 & \textbf{81.8} & 74.1 & 75.6 & 70.2 \\
    \hspace{1.2em}MMStar & \textbf{76.5} & 75.7 & 75.3 & 74.9 & 73.6 \\
    \hspace{1.2em}HallusionBench & 63.5 & 65.2 & \textbf{65.4} & 62.8 & 62.6 \\
    \hspace{1.2em}AI2D & 88.7 & \textbf{89.5} & 84.9 & 86.1 & 88.5 \\
    \hspace{1.2em}MMT-Bench (val) & 70.7 & \textbf{72.7} & 68.1 & 70.9 & 69.7 \\
    \hspace{1.2em}MM-IFEval & 75.7 & \textbf{83.1} & 73.5 & 69.9 & 68.2 \\
    \midrule
    \multicolumn{6}{l}{\textbf{Document \& OCR}} \\
    \hspace{1.2em}OCRBench & 853 & 807 & 819 & 859 & \textbf{879} \\
    \hspace{1.2em}TextVQA (val) & 73.8 & 77.8 & 77.8 & \textbf{80.8} & 79.8 \\
    \hspace{1.2em}DocVQA (val) & 92.8 & 91.3 & \textbf{95.3} & 94.2 & 92.3 \\
    \bottomrule
  \end{tabular}
  }
\end{table}

\subsection{Vision-Language Results}

As shown in Table~\ref{tab:main-results-instruct} and Table~\ref{tab:main-results-thinking}, MiniCPM-o 4.5 demonstrates strong performance across a wide range of vision-language tasks under both instruct and thinking modes.

\paragraph{Comprehensive Capability.}
MiniCPM-o 4.5 achieves an average score of 77.6 on OpenCompass \cite{2023opencompass}, a comprehensive collection of 8 popular vision-language benchmarks, in instruct mode and 78.2 in thinking mode.
With only 9B parameters, it consistently outperforms models of similar scale, such as InternVL3.5-8B \cite{wang2025internvl3} and Qwen3-VL-8B \cite{bai2025qwen3vltechnicalreport}, as well as larger models like Qwen3-Omni-30B \cite{xu2025qwen3omnitechnicalreport}, while close to leading proprietary models including Gemini 2.5 Flash \cite{gemini25} and GPT-5 \cite{singh2025openai}.

\paragraph{OCR and Document Analysis.}
MiniCPM-o 4.5 exhibits the best performance in document parsing. It achieves strong results on OmniDocBench \cite{ouyang2025omnidocbenchbenchmarkingdiversepdf} for both English and Chinese, significantly outperforming other general models with larger parameter size, such as Qwen3-Omni-30B-A3B. On OCRBench \cite{liu2024ocrbench}, TextVQA \cite{singh2019textvqa}, and DocVQA \cite{mathew2021docvqa}, MiniCPM-o 4.5 is on par with top-tier models.

\paragraph{Multi-Image Understanding.}
Benefiting from enhanced data coverage and quality of multi-image datasets, MiniCPM-o 4.5 outperforms all baselines on Mantis-Eval \cite{jiang2024mantis} and MMSI-Bench \cite{yang2025mmsi} as shown in Table~\ref{tab:main-results-instruct}. It also yields a competitive score on MUIRBench \cite{wang2024muirbench}. These results indicate strong performance on cross-image understanding, which is essential for real-world applications.

\begin{table}
  \centering
  \normalsize
  \caption{Results on audio understanding benchmarks. For ASR benchmarks, lower is better; $^{*}$: VoiceBench AlpacaEval scores are rated on a scale from 1 to 5.}
  \label{tab:speech-results}
  \resizebox{0.82\textwidth}{!}{%
  \begin{tabular}{lccc}
    \toprule
    \textbf{Benchmark} & \makecell{\textbf{Kimi-Audio}} & \makecell{\textbf{Qwen3-Omni}} & \makecell{\textbf{MiniCPM-o 4.5}} \\
    \textbf{Size} & \textbf{9B} & \textbf{30B-A3B} & \textbf{9B} \\
    \midrule
    \multicolumn{4}{l}{\textbf{Automatic Speech Recognition}} \\
    \hspace{1.2em}AISHELL-1$\downarrow$ & \textbf{0.6} & \textbf{0.6} & 0.9 \\
    \hspace{1.2em}AISHELL-2$\downarrow$ & 2.6 & \textbf{2.3} & 2.5 \\
    \hspace{1.2em}WenetSpeech test-net$\downarrow$ & 6.3 & \textbf{4.7} & 5.9 \\
    \hspace{1.2em}WenetSpeech test-meeting$\downarrow$ & \textbf{5.4} & 5.9 & 5.7 \\
    \hspace{1.2em}LibriSpeech test-clean$\downarrow$ & 1.3 & \textbf{1.2} & 1.4 \\
    \hspace{1.2em}LibriSpeech test-other$\downarrow$ & \textbf{2.4} & 2.5 & 2.8 \\
    \hspace{1.2em}GigaSpeech test$\downarrow$ & 9.4 & 8.7 & \textbf{8.5} \\
    \hspace{1.2em}VoxPopuli V1-En$\downarrow$ & 8.0 & 6.4 & \textbf{6.2} \\
    \midrule
    \multicolumn{4}{l}{\textbf{Speech Translation}} \\
    \hspace{1.2em}CoVoST 2 en$\rightarrow$zh & 36.6 & 46.6 & \textbf{49.9} \\
    \hspace{1.2em}CoVoST 2 zh$\rightarrow$en & 18.3 & \textbf{29.4} & 26.4 \\
    \midrule
    \multicolumn{4}{l}{\textbf{Multi-task Audio Understanding}} \\
    \hspace{1.2em}MMAU & 68.4 & \textbf{77.5} & 76.9 \\
    \hspace{1.2em}Meld & 59.1 & 56.8 & \textbf{60.2} \\
    \midrule
    \multicolumn{4}{l}{\textbf{Speech Question Answering}} \\
    \hspace{1.2em}VoiceBench AlpacaEval$^{*}$ & 4.46 & 4.74 & \textbf{4.81} \\
    \hspace{1.2em}Speech TriviaQA & 41.9 & 62.9 & \textbf{75.5} \\
    \hspace{1.2em}Speech Web Questions & 46.4 & \textbf{74.9} & 70.2 \\
    \hspace{1.2em}Speech CMMU & \textbf{67.0} & 47.8 & 59.2 \\
    \bottomrule
  \end{tabular}
  }
  \vspace{1mm}  
\end{table}

\subsection{Speech Results}

\paragraph{Audio Understanding.}
As shown in Table~\ref{tab:speech-results}, MiniCPM-o 4.5 demonstrates broad audio understanding capability. 
On ASR, it remains close to the leading systems across both Chinese and English benchmarks, with the best results on GigaSpeech and VoxPopuli. 
More importantly, its advantages extend to semantic speech tasks. MiniCPM-o 4.5 leads on CoVoST 2 en$\to$zh, MELD, VoiceBench AlpacaEval, and Speech TriviaQA, indicating that the model can leverage speech-conditioned representations for translation, audio reasoning, instruction following, and knowledge-intensive speech QA. 
At the same time, the remaining gaps on Speech Web Questions and Speech CMMU show that retrieval-like factual QA and Chinese speech knowledge QA are still challenging.

\paragraph{Speech Generation.}
As shown in Table~\ref{tab:speech-generation-results-restyled}, MiniCPM-o 4.5 demonstrates clear advantages in speech clarity and expressive control. It achieves the lowest CER/WER on SeedTTS Test-ZH and SeedTTS Test-EN, showing reliable bilingual speech generation. On LongTTS, it obtains a much lower English WER than the baselines, indicating better stability for long-form English generation, while remaining close to CosyVoice2 on Chinese CER. It also performs best on Expresso and ESD, suggesting stronger emotion and style control for expressive speech synthesis.

\begin{table}[h]
  \centering
  \footnotesize
  \caption{Speech generation results. Lower is better for CER and WER; N/A: not supported; $^{\ast}$: Neutral reference audio is used for evaluation.}
  \label{tab:speech-generation-results-restyled}
  \resizebox{\linewidth}{!}{%
  \begin{tabular}{lcccccccc}
  \toprule
  \textbf{Model}
  & \multicolumn{2}{c}{\textbf{SeedTTS Test-ZH}}
  & \multicolumn{2}{c}{\textbf{SeedTTS Test-EN}}
  & \multicolumn{2}{c}{\textbf{LongTTS}}
  & \multicolumn{2}{c}{\textbf{Emotion/Style Control}} \\
  \cmidrule(lr){2-3}\cmidrule(lr){4-5}\cmidrule(lr){6-7}\cmidrule(lr){8-9}
  & \textbf{CER}$\downarrow$ & \textbf{SIM-o} & \textbf{WER}$\downarrow$ & \textbf{SIM-o} & \textbf{EN WER}$\downarrow$ & \textbf{ZH CER}$\downarrow$ & \textbf{Expresso}$^{\ast}$ & \textbf{ESD}$^{\ast}$ \\
  \midrule
  CosyVoice2 & 1.45 & \textbf{74.8} & 2.57 & \textbf{65.2} & 14.80 & \textbf{5.27} & 17.9 & 53.4 \\
  Qwen3-Omni & 1.41 & N/A & 3.39 & N/A & 17.33 & 18.99 & N/A & N/A \\
  MiniCPM-o 4.5 & \textbf{0.86} & 74.5 & \textbf{2.38} & 64.9 & \textbf{3.37} & 6.58 & \textbf{29.8} & \textbf{82.1} \\
  \bottomrule
  \end{tabular}
  }
  \end{table}

\subsection{Text Results}

\begin{table}
  \centering
  \caption{Results on text benchmarks.}
  \label{tab:text-reasoning-results-instruct}
  \resizebox{\textwidth}{!}{%
  \begin{tabular}{lccccccccc}
    \toprule
    \textbf{Model} & \textbf{IFEval-PLS} & \textbf{BBH} & \textbf{CMMLU} & \textbf{MMLU} & \textbf{HumanEval} & \textbf{MBPP} & \textbf{Math500} & \textbf{GSM8K} & \textbf{Avg} \\
    \midrule
    Qwen3-8B-Instruct & 83.0 & 69.4 & 78.7 & \textbf{81.7} & \textbf{86.6} & 75.9 & \textbf{84.0} & 93.4 & 81.6 \\
    MiniCPM-o 4.5 & \textbf{84.7} & \textbf{81.1} & \textbf{79.6} & 77.0 & \textbf{86.6} & \textbf{76.7} & 77.0 & \textbf{94.5} & \textbf{82.1} \\
    \bottomrule
  \end{tabular}
  }
\end{table}

As shown in Table~\ref{tab:text-reasoning-results-instruct}, MiniCPM-o 4.5 outperforms its backbone LLM in most text-only tasks, specifically across complex reasoning, mathematics, coding, and instruction following. This suggests that a strategic balance of textual and multimodal data allows the model to retain its text capabilities while acquiring strong multimodal capabilities.

\subsection{Omni-modal and Streaming Results}

\paragraph{Omni-modal Understanding.}
MiniCPM-o 4.5 demonstrates strong omni-modal understanding capabilities as shown in table~\ref{tab:omni-simplex-results}. It achieves the best results on five of the seven benchmarks, namely Daily-Omni, WorldSense, Video-Holmes, JointAVBench, and AVUT-Human. Despite its small parameter-size, it remains competitive on FutureOmni and Video-MME-Short (w/ audio).

\begin{table}[h]
  \centering
  \footnotesize
  \caption{Omni-modal benchmark results in simplex settings.}
  \label{tab:omni-simplex-results}
  \setlength{\tabcolsep}{4pt}
  \begin{tabular}{@{}lccc@{}}
  \toprule
  \textbf{Benchmark} & \makecell{\textbf{Gemini 2.5 Flash}} & \makecell{\textbf{Qwen3-Omni}} & \textbf{MiniCPM-o 4.5} \\
  \textbf{Size} & \textbf{-} & \textbf{30B-A3B} & \textbf{9B} \\
  \midrule
  Daily-Omni & 79.3 & 70.7 & \textbf{80.2} \\
  WorldSense & 52.6 & 54.0 & \textbf{55.7} \\
  Video-Holmes & 51.3 & 50.4 & \textbf{64.3} \\
  JointAVBench & 55.6 & 53.1 & \textbf{60.0} \\
  AVUT-Human & 65.4 & 74.2 & \textbf{78.6} \\
  FutureOmni & 55.6 & \textbf{62.1} & 56.1 \\
  Video-MME-Short (w/ audio) & \textbf{85.5} & 81.3 &
  84.7 \\
  \bottomrule
  \end{tabular}
\end{table}

\paragraph{Full-Duplex Results.}

Table~\ref{tab:omni-duplex-results} evaluates whether models can respond appropriately while continuously receiving visual streams. MiniCPM-o 4.5 achieves a win rate of $54.4$ on LiveSports-3K-CC, outperforming LiveCC and StreamingVLM by $12.9$ and $8.8$ points, respectively. This improvement suggests that Omni-Flow is effective for continuous visual interaction: by organizing perception and response along a shared timeline, the model can better ground its responses in the evolving scene instead of relying on delayed or fragmented visual context.

\begin{table}[h]
\vspace*{-4mm}
  \centering
  \footnotesize
  \caption{Vision-only full-duplex benchmark results.}
  \label{tab:omni-duplex-results}
  \setlength{\tabcolsep}{4pt}
  \begin{tabular}{@{}lccc@{}}
  \toprule
  \textbf{Benchmark} & \makecell{\textbf{LiveCC}} & \textbf{StreamingVLM} & \textbf{MiniCPM-o 4.5} \\
  \textbf{Size} & \textbf{8B} & \textbf{8B} & \textbf{9B} \\
  \midrule
  LiveSports-3K-CC & 41.5 & 45.6 & \textbf{54.4} \\
  \bottomrule
\vspace*{-4mm}  \end{tabular}
  \end{table}

  \subsection{Analysis}

  \begin{figure}[t]
    \centering
    \vspace{0em}
  
  \begin{minipage}[t]{0.6\textwidth}
    \vspace{0pt}
    \centering
    \captionof{table}{Performance of different Length reward strategies.}
    \label{tab:length-penalty-ablation}
    \vspace{0.6em}
  
    \renewcommand{\arraystretch}{1.25}
    \setlength{\tabcolsep}{5pt}
  
    \resizebox{\linewidth}{!}{%
    \begin{tabular}{lcccc}
      \toprule
      \multirow{2}{*}{\textbf{Length Reward}} 
      & \multicolumn{2}{c}{\textbf{Benchmarks Avg.}} 
      & \multicolumn{2}{c}{\textbf{Length Reduction Avg.}} \\
      \cmidrule(lr){2-3} \cmidrule(lr){4-5}
      & \textbf{Thinking} 
      & \textbf{Instruct} 
      & \textbf{Thinking} 
      & \textbf{Instruct} \\
      \midrule
      No Length Reward
      & 73.5 & \textbf{70.9} & -- & -- \\
      Kimi K1.5-Style~\cite{team2025kimi} 
      & 73.0 & 70.1 & \textbf{50.7\%} & 20.2\% \\
      Ours
      & \textbf{74.3} & \textbf{70.9} & 35.3\% & \textbf{20.5\%} \\
      \bottomrule
    \end{tabular}
    }
  
    \renewcommand{\arraystretch}{1.0}
  \end{minipage}
    \hfill
    \begin{minipage}[t]{0.39\textwidth}
      \vspace{0pt}
      \centering
      \captionof{figure}{Training set accuracy using different length penalty methods.}
      \includegraphics[width=\linewidth]{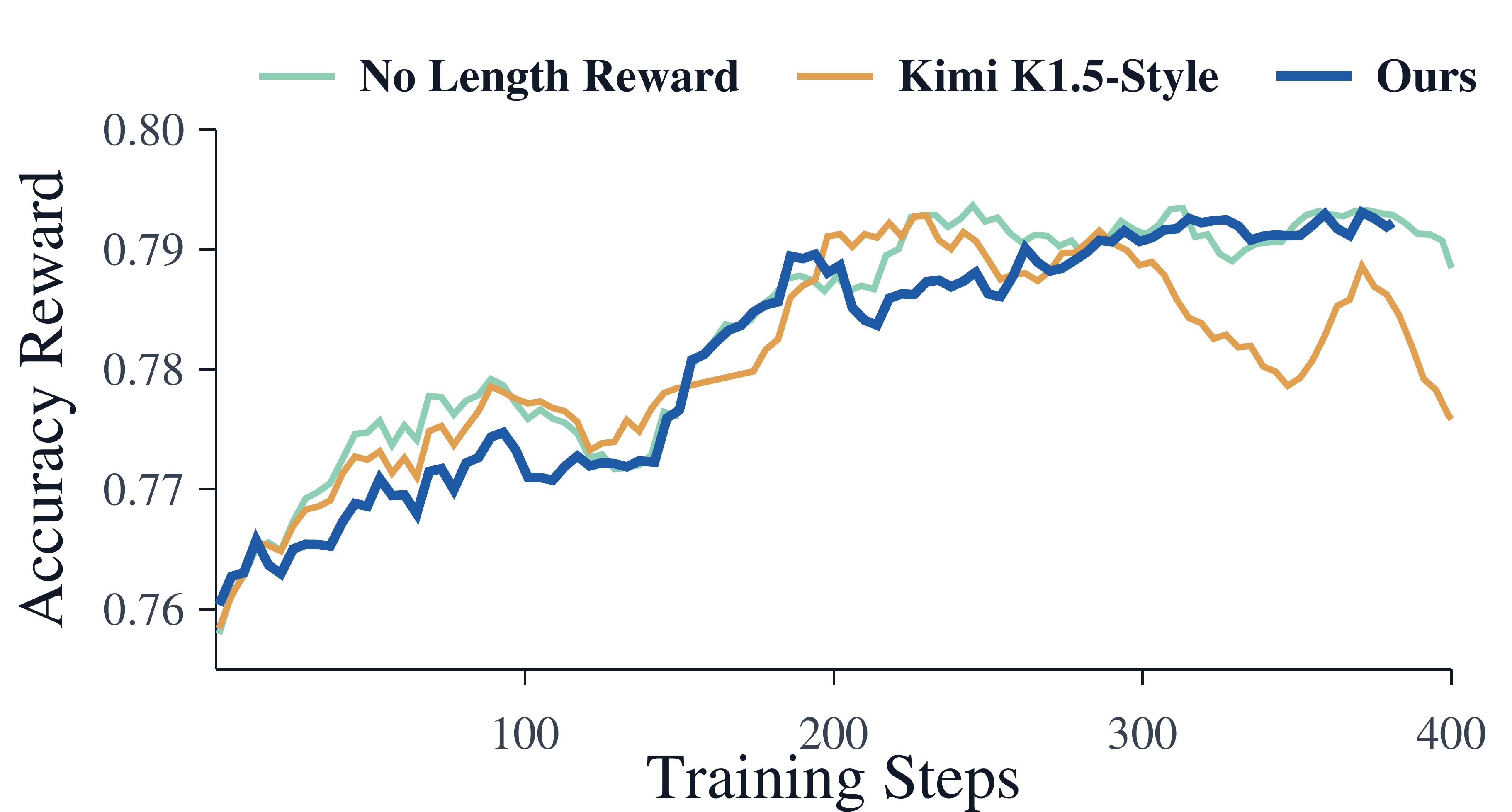}
      \label{fig:training-set-accuracy}
    \end{minipage}
  
    \vspace{-0.9em}
\end{figure}

\paragraph{Ablation of Length Reward.}
We ablate the length reward design to examine the trade-off between response efficiency and task performance. We conduct a lightweight RL training experiment and report average results on MMBench, MathVista, MMMU, AI2D, OCRBench, HallusionBench and MMStar.
We compare the Kimi K1.5-style length reward~\cite{team2025kimi} with our proposed smooth length reward. 
As shown in Table~\ref{tab:length-penalty-ablation}, the K1.5-style reward aggressively reduces the response length in thinking mode by $50.7\%$, but also decreases the benchmark average from $73.5$ to $73.0$. 
In contrast, our method achieves a more moderate length reduction of $35.3\%$ on thinking tasks, while improving the benchmark average to $74.3$. 
For instruction mode, both methods reduce the response length by around $20\%$, while our method maintains the best average performance.
The training curves in Figure~\ref{fig:training-set-accuracy} further explain the difference between these designs. 
The K1.5-style reward shows a clear slowdown and even slight degradation in training accuracy in the later stage, suggesting that an overly aggressive length reward can conflict with the accuracy reward and suppress further optimization. 
Our method avoids this instability through smoother reward shaping, maintaining a training trajectory closer to the baseline without length reward while still achieving substantial length reduction. 
These results indicate that our length reward provides a better efficiency-performance trade-off: it removes unnecessary long reasoning without overly penalizing useful intermediate reasoning steps.

\begin{wraptable}{r}{0.5\textwidth}
  \centering
  \vspace{-4mm}
  \small
  \caption{
  MiniCPM-o 4.5 speech generation quality of different modes. We report results on Seed TTS test set.
  }
  \label{tab:streaming-speech-ablation}
  \resizebox{0.48\textwidth}{!}{%
  \begin{tabular}{lcccc}
    \toprule
    \textbf{Interleaving Mode} 
    & \makecell{\textbf{ZH}\\\textbf{CER}$\downarrow$}
    & \makecell{\textbf{ZH}\\\textbf{SIM-o}$\uparrow$}
    & \makecell{\textbf{EN}\\\textbf{WER}$\downarrow$}
    & \makecell{\textbf{EN}\\\textbf{SIM-o}$\uparrow$} \\
    \midrule
    No interleave
    & 1.44 & 74.1 & 2.70 & 64.9 \\
    Fixed text
    & \textbf{0.86} & \textbf{74.5} & \textbf{2.38} & 64.9 \\
    Dynamic text (TAIL)
    & 1.04 & 74.1 & 3.93 & \textbf{65.1} \\
    \bottomrule
  \end{tabular}
  }
  \vspace{-3mm}
\end{wraptable}

\paragraph{Comparison of Speech Generation Modes.}
Table~\ref{tab:streaming-speech-ablation} compares three speech generation modes: non-interleaved generation, our fixed-text interleaving, and our dynamic-text interleaving strategy TAIL. Fixed-text interleaving achieves the best CER/WER, suggesting that chunked streaming generation can improve pronunciation accuracy over synthesizing speech after the full text is generated. TAIL is designed for the more challenging full-duplex setting, where text and speech must stay temporally aligned. Although it slightly sacrifices recognition accuracy, especially on English WER, it maintains reasonable overall speech quality, hitting a practical trade-off between streaming interaction and speech generation quality.
\section{Efficient Real-Time Inference}

We first evaluate the inference efficiency of MiniCPM-o 4.5 under the standard vLLM~\cite{kwon2023efficientmemorymanagementlarge} setting. 
As shown in Table~\ref{tab:vllm_inference_efficiency}, compared with Qwen3-Omni-30B-A3B, MiniCPM-o 4.5 shows clear advantages in both throughput and memory usage on a single NVIDIA RTX 4090. 
In BF16, Qwen3-Omni-30B-A3B runs out of memory, while MiniCPM-o 4.5 achieves 154.3 tokens/s with 19 GB memory usage. 
In INT4, MiniCPM-o 4.5 further achieves 212.3 tokens/s, lower first-token latency, and nearly half the memory usage compared with Qwen3-Omni-30B-A3B.

To further improve deployment efficiency for the full-duplex streaming mode, we develop an efficient inference framework based on llama.cpp~\cite{ggml2023llamacpp}, termed \textit{llama.cpp-omni}. 
The framework is tailored to the streaming interaction paradigm of MiniCPM-o 4.5 and enables smooth execution across multiple hardware platforms. 
Beyond runtime efficiency, we also validate its compatibility across different operating systems, including macOS, Windows, and Linux. 
We further provide a lightweight demo system, allowing users to quickly deploy MiniCPM-o 4.5 on their own hardware and experience its real-time speech, vision-language, and full-duplex omni-modal interaction capabilities.
Table~\ref{tab:framework_inference_efficiency} compares the real-time factor (RTF) and memory usage of different inference frameworks across hardware configurations. 
Compared with the PyTorch implementation, \textit{llama.cpp-omni} substantially reduces RTF on both RTX 4090 and DGX Spark while maintaining a lower memory footprint under INT4 quantization, demonstrating its effectiveness for efficient real-time deployment.

\begin{table}[t]
    \centering
    \caption{
    Inference efficiency comparison between MiniCPM-o 4.5 and Qwen3-Omni-30B-A3B on a single NVIDIA RTX 4090 using vLLM. 
    First-token latency is evaluated with 64-frame visual inputs, while throughput and memory usage are measured on text-only tasks. 
    OOM denotes out-of-memory.
    }
    \label{tab:vllm_inference_efficiency}
    \resizebox{0.82\textwidth}{!}{
    \begin{tabular}{llccc}
        \toprule
        \textbf{Model} 
        & \textbf{Dtype}
        & \textbf{Throughput} $\uparrow$
        & \textbf{First-token Latency} $\downarrow$
        & \textbf{Memory} $\downarrow$ \\
        & 
        & \textbf{(tokens/s)}
        & \textbf{(s)}
        & \textbf{(GB)} \\
        \midrule
        Qwen3-Omni-30B-A3B & BF16 & OOM & OOM & OOM \\
        MiniCPM-o 4.5      & BF16 & 154.3 & 0.59 & 19 \\
        \midrule
        Qwen3-Omni-30B-A3B & INT4 & 147.8 & 0.98 & 20 \\
        MiniCPM-o 4.5      & INT4 & \textbf{212.3} & \textbf{0.58} & \textbf{11} \\
        \bottomrule
    \end{tabular}
    }
\end{table}

\begin{table}[t]
    \centering
    \caption{
    Inference efficiency comparison of different inference frameworks for MiniCPM-o 4.5. 
    We report the real-time factor (RTF) and memory usage on different hardware configurations. 
    Lower RTF indicates higher inference efficiency. 
    OOM denotes out-of-memory.
    }
    \label{tab:framework_inference_efficiency}
    \resizebox{0.78\textwidth}{!}{
    \begin{tabular}{llcccc}
        \toprule
        \multirow{2}{*}{\textbf{Framework}} 
        & \multirow{2}{*}{\textbf{Dtype}}
        & \multicolumn{2}{c}{\textbf{RTX 4090}}
        & \multicolumn{2}{c}{\textbf{DGX Spark}} \\
        \cmidrule(lr){3-4} \cmidrule(lr){5-6}
        & 
        & \textbf{RTF} $\downarrow$ 
        & \textbf{Memory (GB)} $\downarrow$
        & \textbf{RTF} $\downarrow$ 
        & \textbf{Memory (GB)} $\downarrow$ \\
        \midrule
        PyTorch & BF16 & OOM & OOM & 2.43 & 26 \\
        PyTorch & INT4   & 1.26 & 14  & 1.27 & 14 \\
        llama.cpp-omni (Ours) & FP16 & 0.27 & 19 & 0.46 & 19 \\
        llama.cpp-omni (Ours) & INT4 & \textbf{0.21} & \textbf{11} & \textbf{0.20} & \textbf{11} \\
        \bottomrule
    \end{tabular}
    }
\end{table}

\section{Conclusion}

\textbf{Contributions.}
We present MiniCPM-o 4.5, a 9B open-source MLLM for real-time full-duplex omni-modal interaction. By continuously perceiving visual and auditory streams while generating speech responses, MiniCPM-o 4.5 moves beyond conventional turn-based multimodal interaction and enables a more human-like interaction paradigm. It achieves this capability with practical edge efficiency, requiring less than 12GB RAM during deployment, while also approaching Gemini 2.5 Flash in vision-language capabilities and delivering frontier image and video understanding performance among open-source MLLMs at this scale. We further introduce the unified omni-modal streaming framework Omni-Flow, as the key technique behind MiniCPM-o 4.5, that aligns multimodal inputs and outputs along a shared temporal axis, providing a general formulation for full-duplex and proactive multimodal interaction.

\textbf{Limitations.}
MiniCPM-o 4.5 is still an early exploration of real-time full-duplex omni-modal interaction and remains limited in several aspects. First, its foundation capability and robustness in long, dynamic real-world streaming interactions still require further improvement and validation. Second, speech generation in omni-modal streaming mode can occasionally be unstable, including mispronunciation or unintended mixing between English and Chinese. Third, although our web demo enables convenient access, users may experience increased latency or missing output fragments under unstable network conditions; local deployment with \textit{llama.cpp-omni} can better support smooth real-time interaction. Finally, the model's proactive behavior is still relatively simple, leaving richer context-aware planning and self-initiated assistance for future work.

{
\bibliographystyle{unsrt}
}
\bibliography{neurips_2024}

\section{Appendix}

\appendix

\section{Model Configuration}
\label{app:config}

Table~\ref{tab:model_config} lists the architectural hyperparameters of each component. The full model contains 9.34B learnable parameters and uses bfloat16 precision.

\begin{table}[h]
\centering
\caption{Architectural hyperparameters of MiniCPM-o 4.5.}
\label{tab:model_config}
\small
\begin{tabular}{lll}
\toprule
\textbf{Component} & \textbf{Hyperparameter} & \textbf{Value} \\
\midrule
\multicolumn{3}{l}{\textbf{Visual Encoder} (SigLIP ViT, 417.8M)} \\
& Hidden dimension          & 1,152 \\
& Layers                    & 27 \\
& Attention heads           & 16 \\
& FFN dimension             & 4,304 \\
& Activation                & GELU$_{\text{tanh}}$ \\
& Patch size                & $14 \times 14$ \\
\midrule
\multicolumn{3}{l}{\textbf{Visual Resampler} (88.9M)} \\
& Query tokens              & 64 \\
& Embedding dimension       & 4,096 \\
& Attention heads           & 32 \\
\midrule
\multicolumn{3}{l}{\textbf{Audio Encoder} (Whisper Medium encoder, 307.2M)} \\
& Hidden dimension          & 1,024 \\
& Layers                    & 24 \\
& Attention heads           & 16 \\
& FFN dimension             & 4,096 \\
& Activation                & GELU \\
& Mel-frequency bins        & 80 \\
\midrule
\multicolumn{3}{l}{\textbf{Audio Projector} (21.0M)} \\
& Architecture              & Two-layer MLP with ReLU \\
& Dimensions                & $1024 \to 4096 \to 4096$ \\
\midrule
\multicolumn{3}{l}{\textbf{LLM Backbone} (Qwen3-8B, 8,189.2M)} \\
& Hidden dimension          & 4,096 \\
& Layers                    & 36 \\
& Attention heads           & 32 \\
& KV heads (GQA)            & 8 \\
& Head dimension            & 128 \\
& FFN dimension             & 12,288 \\
& Activation                & SiLU \\
& Normalization             & RMSNorm ($\epsilon{=}10^{-6}$) \\
& Vocabulary size           & 151,748 \\
& Max context length        & 40,960 \\
& RoPE $\theta$             & $10^{6}$ \\
& Weight tying              & None \\
\midrule
\multicolumn{3}{l}{\textbf{Backbone-to-Decoder Projector} (10.5M)} \\
& Architecture              & Two-layer MLP with ReLU \\
& Dimensions                & $4096 \to 768 \to 768$ \\
\midrule
\multicolumn{3}{l}{\textbf{Speech Token Decoder}} \\
& Text embedding layer            & 116.8M \\
& Text vocabulary size      & 152,064 \\

& Transformer                & 188.8M \\
& Hidden dimension          & 768 \\
& Layers                    & 20 \\
& Attention heads           & 12 \\
& KV heads                  & 12 \\
& FFN dimension             & 3,072 \\
& Activation                & SiLU \\
& Max context length        & 4,096 \\

& Speech codebook size      & 6,562 \\
& Speech number of codebooks       & 1 \\
& Speech token frame rate       & 25/s\\
\bottomrule
\end{tabular}
\end{table}


\end{document}